\documentclass[runningheads]{llncs}

\usepackage{algorithm}
\usepackage{algpseudocode}
\usepackage{booktabs}
\usepackage{multirow}
\usepackage{mathtools}
\usepackage{graphicx}
\usepackage{accv}
\usepackage{hyperref}
\usepackage{orcidlink}

\begin{document}

\title{Drift-adaptive Representation Preservation for Domain-Incremental Learning}

\titlerunning{Drift-adaptive Representation}

\author{Hyunmin Kong\orcidlink{0009-0007-4649-6522} \and
Gyutae Oh\orcidlink{0009-0008-6774-1539} \and
Jitae Shin\thanks{Corresponding author.}\orcidlink{0000-0002-2599-3331}}

\authorrunning{Hyunmin Kong et al.}

\institute{Department of Electrical and Computer Engineering,\\
Sungkyunkwan University,\\
Suwon 16419, Republic of Korea\\
\email{\{khm1754, alswo740012, jtshin\}@skku.edu}}

\maketitle

\begin{abstract}

Pretrained vision models provide strong general-purpose representations, but in Domain-Incremental Learning (DIL), where the input distribution changes
sequentially, adapting to new domains often degrades performance on previously
learned ones. We observe that forgetting is not uniform across domains and classes and is not explained by training order alone. This motivates us to track each past domain--class concept individually. We use representation drift, defined as the deviation of a past concept's current representation from a fixed anchor, as a measurable surrogate for forgetting risk. The measured drift then serves two complementary roles. A consolidation loss assigns stronger preservation to high-drift pairs, while prototype construction down-weights highly drifted domains to reduce their influence on the class reference. Preservation is thus
allocated by measured drift rather than assumed importance. To obtain this signal during training without storing past raw data, we repurpose a Joint-Embedding Predictive Architecture (JEPA). For each domain--class pair, we store only a compact context summary in the backbone feature space and rerun it through the learned context-to-target relation to measure its drift from the fixed anchor. Across DomainNet, Office-Home, CORe50, and
CDDB-Hard, our method sustains a strong accuracy--forgetting balance on every
benchmark, using substantially less training memory than accuracy-competitive
methods and the smallest storage among compared methods.

\end{abstract}

\begin{figure}[t]
\centering
\includegraphics[width=0.7\textwidth]{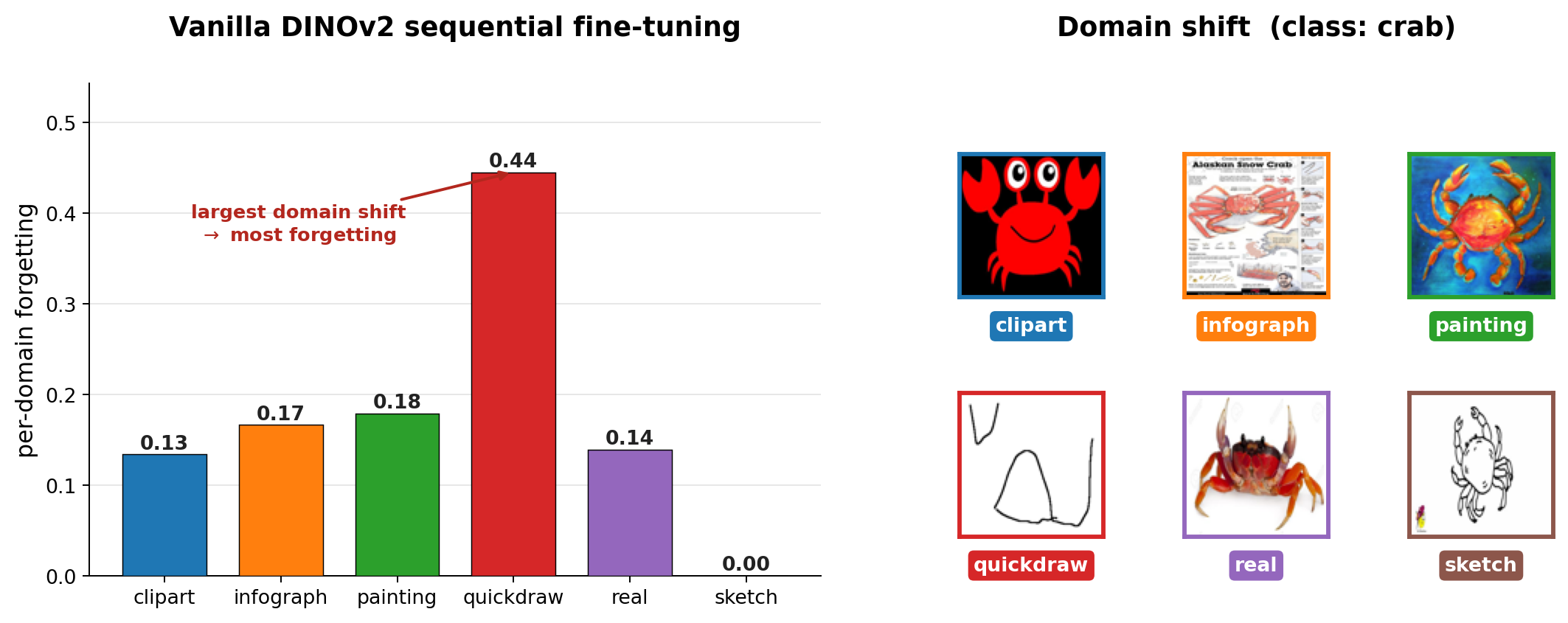}
\caption{Per-domain forgetting when DINOv2 is sequentially fine-tuned on DomainNet. Example images of the same class (e.g., crab) illustrate the visual variation across domains.} \label{fig:perdomain}
\end{figure}

\section{Introduction}
\label{sec:intro}

Pretrained vision models provide strong, broadly transferable representations and have substantially expanded the practical promise of DIL~\cite{wang2024non} \cite{zhou2025dual}. In DIL, a model learns a sequence of domains that share a label space but differ in input distribution, and it must maintain stable performance on all domains seen so far without revisiting past data~\cite{wang2025dualcp} \cite{wang2025boosting}. The central difficulty is the stability--plasticity dilemma: adapting the representation to a new domain disturbs the structure formed on past domains and induces forgetting, whereas suppressing representational change too strongly limits adaptation \cite{liang2024inflora} \cite{xie2022general}.

We study forgetting in DIL through the lens of representation drift. As the model learns a domain, the resulting per-class representation serves as a fixed reference point, or anchor, against which later changes can be measured. When the model subsequently adapts to new domains, a previously learned concept may move away from this reference. We use this deviation as a measurable signal of how strongly the concept has changed and as a surrogate for its forgetting risk. Crucially, these changes are not uniform across previously learned domains and classes. As shown in Fig.~\ref{fig:perdomain}, the amount of forgetting varies substantially across domains within the same training sequence and cannot be characterized by training order alone. This heterogeneous behavior motivates us to track representation drift separately for each domain--class pair rather than relying on a single global measure. This nonuniformity creates a specific problem for prototype-based classification: a pair with large drift has likely departed from the stable class semantics, and combining it with equal weight when forming the class prototype lets the distorted representation contaminate the class reference itself---an effect we call prototype contamination.

Motivated by this, we use drift in two ways at once. First, we use drift as a proxy for forgetting risk: a pair with larger drift receives stronger preservation because its current reconstruction is less aligned with the fixed class anchor. Second, drift is a reliability signal for prototype construction: a pair with small drift is trusted, whereas a pair with large drift is down-weighted, shielding the class reference from contamination. A single measured drift signal thus plays two complementary roles, so that preservation is allocated according to measured drift rather than assumed importance.

Using drift this way has a prerequisite: it must be measurable at the domain--class level, during training, and without access to past raw data. We
observe that JEPA \cite{assran2023self} provides a natural mechanism for this purpose. Because it learns, in latent space, to predict one observed cue from another, a past concept can be re-examined by storing only a compact context summary for each domain--class pair and rerunning it through this learned relation. Concretely, a frozen backbone's CLS-to-patch attention splits each image into a target carrying class-relevant evidence and a context capturing domain-sensitive cues; the model learns to predict the target from the context, and stored past-domain context
summaries are re-input to the current encoder and predictor to measure how far the
prediction departs from the fixed anchor---this distance is the drift signal.

This design also brings a concrete efficiency benefit: no gradient flows through the frozen backbone, so training is confined to a lightweight encoder and predictor and the training memory footprint remains small, while past information is stored only as compact per-domain--class summaries, so storage grows only mildly with the number of domains. Our contributions are summarized as follows:
\begin{itemize}
  \item We propose a preservation framework in which a single measured
  drift signal drives two complementary mechanisms---a drift-adaptive
  consolidation loss that concentrates preservation on high-drift concepts, and
  a robust prototype construction that down-weights high-drift domains to
  prevent prototype contamination---allocating preservation by measured drift
  rather than assumed importance.
  \item To make this signal available at the domain--class level during
  training, without storing any past image, we repurpose JEPA's learned
  context-to-target relation as a drift meter, rerunning a stored domain--class
  context summary through the current encoder and predictor.
  \item Across four DIL benchmarks, our method sustains a strong
  accuracy--forgetting balance on every benchmark while using substantially less
  training memory than accuracy-competitive methods and the smallest storage
  among the compared methods.
\end{itemize}

\section{Related Work}
\label{sec:related}

\subsection{Domain-Incremental Learning}
\label{sec:rw-dil}

Continual learning studies how a model can absorb new data distributions without
forgetting previously learned ones. In domain-incremental learning, all domains
share the same label space while the input distribution shifts sequentially, so
the central difficulty is to keep past domains from degrading as the
representation adapts to new ones. Rehearsal-based methods address this by storing
or synthesizing past samples for replay~\cite{rebuffi2017icarl} \cite{buzzega2020dark}. They
are effective, but their storage cost grows with the data, they raise privacy
concerns, and they are inapplicable when past images cannot be retained \cite{zhuang2024gacl} \cite{gomez2024exemplar}.
Regularization-based methods instead constrain parameters or outputs deemed
important for earlier tasks from changing substantially, as in EWC and
LwF~\cite{li2017learning} \cite{kirkpatrick2017overcoming}. Such preservation, however, is driven by
an importance that is assumed or estimated globally, rather than by measuring
which domain--class concepts are actually drifting under the current model. Recent domain-incremental and pretrained-model-based continual learning methods
reduce reliance on raw data replay by keeping the pretrained backbone frozen and
adapting to new domains through a small number of additional parameters or memory.
Prompt-based methods such as CODA-Prompt \cite{smith2023coda}, OS-Prompt \cite{kim2024one}, and OS-Prompt++ \cite{kim2024one} adapt by
accumulating or selecting prompt parameters per
domain. In contrast, DIDD \cite{oh2026asymmetric},
GraphKeeper \cite{guo2026graphkeeper}, SOYO \cite{wang2025boosting}, DUCT \cite{zhou2025dual}, and MoE \cite{yu2024boosting} preserve or route domain knowledge
through different mechanisms---feature- and graph-based memories, selective
routing over domain-specific parameters or paths, backbone adaptation, and expert
modules. Our method instead keeps the
pretrained backbone frozen and stores only a compact summary per domain--class
pair, using measured drift to decide what to preserve and which contributions to
trust when forming each prototype.

\subsection{Representation Drift and Predictive Representations}
\label{sec:rw-drift}

The instability of learned representations under sequential training is a
well-recognized difficulty in continual learning. As a model adapts to new data,
the features of previously learned concepts may move, causing stored statistics
such as class prototypes to fall out of alignment with the current feature
space---an effect often described as representation, feature, or semantic
drift~\cite{li2024fcs} \cite{goswami2024resurrecting} \cite{gomez2024exemplar}. Prior work has mainly treated this drift as an error to
be compensated for or suppressed, for example by estimating the displacement of
stored prototypes or by stabilizing old representations through feature alignment
and distillation~\cite{magistri2024elastic}. In contrast, we use drift before suppressing
it: we measure it at the domain--class level and use its magnitude as a
nonuniform signal that determines both how strongly each concept should be
preserved and how much each domain should contribute to the class prototype. Our drift measurement is enabled by a latent predictive relation inspired by
JEPA, which learns representations by predicting the embedding of one part or
view of an input from another in latent space, typically using an online encoder
and an exponential moving average target encoder. In their
original use, this predictive objective is designed for self-supervised
representation learning. We repurpose the same principle for continual learning:
the context-to-target relation trained on the current domain is reapplied to
stored past-domain context summaries, and the predicted targets are compared with
fixed class anchors to measure drift. Measurement and learning therefore share one
function rather than relying on a separate probe, and the resulting drift signal
drives both drift-adaptive consolidation and robust prototype construction. This
use of a predictive relation as a per-concept meter of drift, with a single signal
serving both roles, sets our approach apart from prior drift compensation and
pretrained-model-based continual learning methods.

\begin{figure}[t]
\centering
\includegraphics[width=0.7\textwidth]{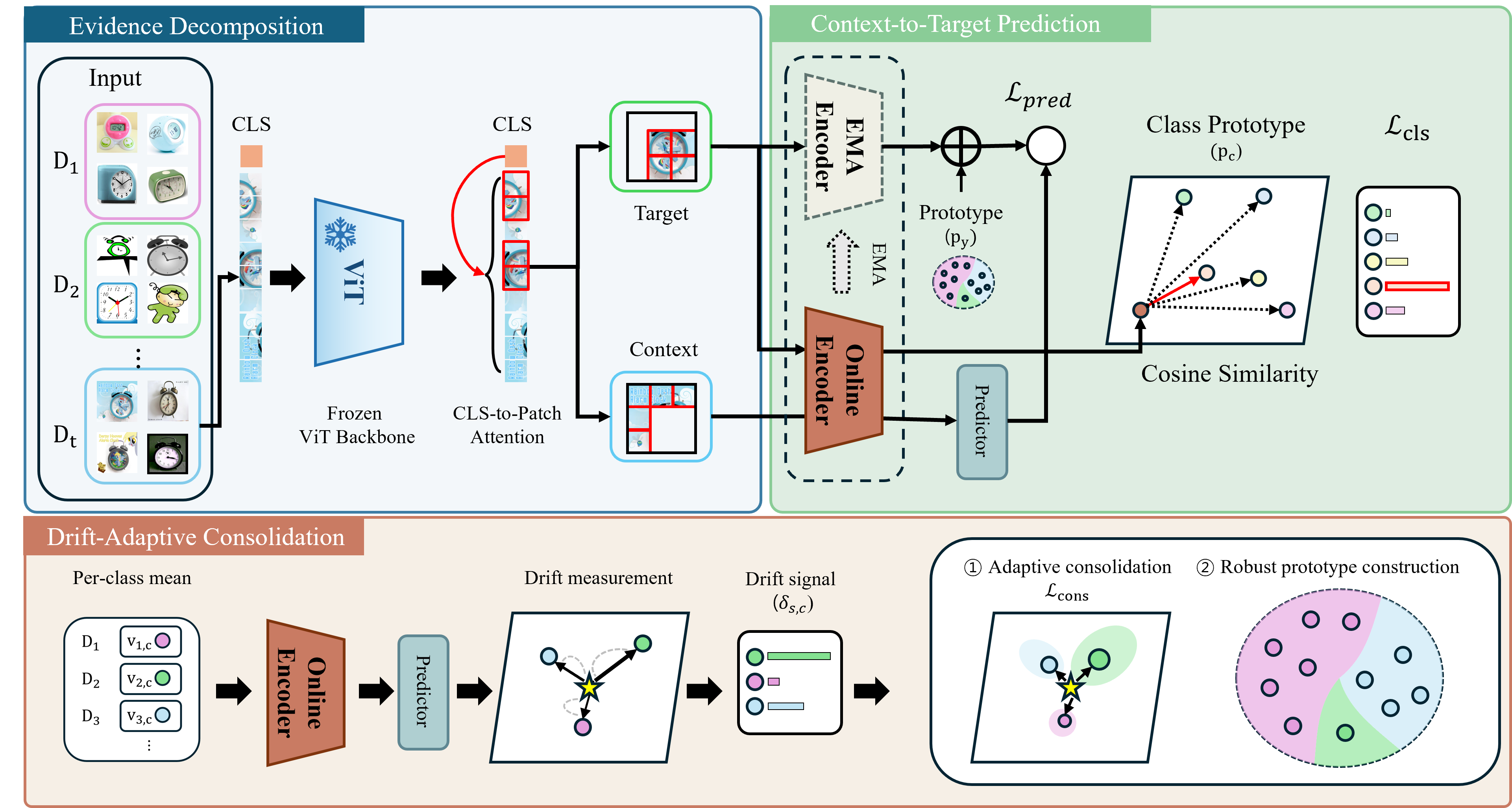}
\caption{Overview of our framework. Evidence Decomposition splits each input
image into a target and a context. Context-to-Target Prediction predicts the
target from the context and performs classification. Drift-Adaptive Consolidation
measures the drift from the fixed anchors using stored past-domain summaries
while learning a new domain, and uses this signal to modulate the preservation
strength and to weight each domain's contribution when forming the class
prototypes.} \label{fig:overview}
\end{figure}

\section{Proposed Method}

\subsection{Problem Formulation}
\label{sec:problem-setup}

In DIL, a model is trained on $T$ domains in sequence. Each domain
$t \in \{1, \dots, T\}$ is associated with a distribution $\mathcal{P}_t(X, Y)$,
from which a training set $\mathcal{D}_t = \{(x_i^t, y_i^t)\}_{i=1}^{N_t}$ is
sampled, where $x_i^t$ is an input image, $y_i^t \in \mathcal{Y}$ is its label,
and $N_t$ is the number of samples in domain $t$. All domains share a common
label space $\mathcal{Y} = \{1, \dots, C\}$ with $C$ classes, while the input
distribution $\mathcal{P}_t(X)$ differs across domains.

Training proceeds sequentially, one domain at a time: at step $t$ the model can
access only the current domain $\mathcal{D}_t$ and cannot use the original images
of the previous domains $\mathcal{D}_1, \dots, \mathcal{D}_{t-1}$. Let $\Phi_t$
denote the model trained up to domain $t$. The objective is to maximize the
average performance over all domains seen so far,
\begin{equation}
  A(\Phi_t) \;=\; \frac{1}{t}\sum_{s=1}^{t}
  \mathbb{E}_{(x,y)\sim \mathcal{P}_s}
  \big[\,\mathbf{1}(\Phi_t(x)=y)\,\big],
  \label{eq:dil-objective}
\end{equation}
where $\mathbf{1}(\cdot)$ is the indicator function and $s \le t$ indexes the
domains seen so far. Keeping the objective in Eq.~\eqref{eq:dil-objective} without access to past data is the central challenge of DIL.

In this work, $\Phi_t$ is a prototype-based classifier that maps an input to a
representation through a frozen pretrained backbone and a trainable encoder, and
classifies it by the cosine similarity to each class prototype. Under this formulation, forgetting is reflected in the degradation of performance on previously learned domains. We track the underlying representational change through the deviation of each past concept from its fixed class anchor and use this representation drift as a measurable surrogate for forgetting risk. At
each domain transition, before training on domain $t$, we fix the representation
of each class observed so far as a reference against which subsequent change is
measured, and call it the anchor $a_c^{(t-1)}$ (constructed in
\S\ref{sec:anchor}). As the encoder changes while later domains are learned, a past
concept may be mapped to a location that departs from its anchor, and this
distance serves as a measurable surrogate for the concept's forgetting risk.

To make this distance concrete, we define the drift of class $c$ from domain $s$,
measured at step $t$, as a nonnegative scalar $\delta_{s,c}^{(t)}$. This construction has two important aspects. First, $\delta_{s,c}^{(t)}$ is obtained without any
past image: a past concept is reconstructed through the current model from a
stored domain--class context summary and compared against its anchor. Second, since the anchor is
fixed per class, the per-domain difference in drift arises solely from the
reconstructed past representation; drift is therefore not a single global quantity
but one indexed by each domain--class pair $(s,c)$. We track these domain--class-specific representational changes through the per-pair surrogate $\delta_{s,c}^{(t)}$ during training, and use this signal to allocate preservation and construct robust prototypes.

\subsection{Evidence Decomposition and Prediction}
\label{sec:evidence-prediction}

We now specify the representation used for classification, introduced abstractly
in \S\ref{sec:problem-setup}, as two complementary cues derived from attention,
and learn a predictive relation between them. This predictive relation later
serves, in \S\ref{sec:drift-measure}, as the function that measures the drift of past
concepts.

\subsubsection{CLS-to-Patch Attention}
\label{sec:cls-attn}

Passing an input $x$ through the frozen backbone $f(\cdot)$ yields a CLS token and
$N$ patch tokens. Let $h_n \in \mathbb{R}^{D}$ be the $n$-th patch token, where $D$
is the backbone token dimensionality, and let $\alpha_n$ be the attention from the
CLS token to that patch, normalized over the patches so that $\alpha_n \geq 0$ and
$\sum_n \alpha_n = 1$; intuitively, $\alpha_n$ reflects the regions that the
pretrained backbone attends to more strongly and that therefore tend to carry the
class-relevant evidence \cite{darcet2024vision}. Using the attention weights $\alpha_n$ and their
complementary weights $\tilde{\alpha}_n$, we aggregate the patch tokens into two
representations,
\begin{equation}
  \bar{u} = \sum_{n=1}^{N} \alpha_n\, h_n,
  \qquad
  \bar{v} = \sum_{n=1}^{N} \tilde{\alpha}_n\, h_n,
  \qquad
  \tilde{\alpha}_n = \frac{1-\alpha_n}{\sum_{m=1}^{N}(1-\alpha_m)},
  \label{eq:pool}
\end{equation}
which we then normalize to unit vectors,
\begin{equation}
  u^{\mathrm{raw}} = \frac{\bar{u}}{\lVert \bar{u} \rVert_2},
  \qquad
  v^{\mathrm{raw}} = \frac{\bar{v}}{\lVert \bar{v} \rVert_2},
  \qquad
  u^{\mathrm{raw}}, v^{\mathrm{raw}} \in \mathbb{R}^{D},
  \label{eq:pool-normalize}
\end{equation}
where $u^{\mathrm{raw}}$ is the target, formed from regions emphasized by
the pretrained backbone and used as the classification cue, whereas
$v^{\mathrm{raw}}$ is the context, formed with the complementary weights
$\tilde{\alpha}_n$ and tending to capture cues sensitive to domain change, such as
background, style, and texture.

\subsubsection{Latent Encoding and Classification}
\label{sec:encode-classify}

The two raw representations are passed through a shared trainable encoder
$g_\theta(\cdot)$ and mapped to the embedding space $\mathbb{R}^{d}$:
\begin{equation}
  u \;=\; g_\theta(u^{\mathrm{raw}}),
  \qquad
  v \;=\; g_\theta(v^{\mathrm{raw}}),
  \qquad u, v \in \mathbb{R}^{d},
  \label{eq:encode}
\end{equation}
where $d$ is the encoder output dimensionality, $u$ is the target representation
used for classification, and $v$ is the context representation used as the input
to prediction; the same embedding space is later used for drift measurement.
Throughout, all encoder and predictor outputs are $\ell_2$-normalized to unit
vectors before any cosine comparison. As described in \S\ref{sec:problem-setup},
classification is performed via the cosine similarity between the target
representation $u$ and the class prototype $p_c \in \mathbb{R}^{d}$,
\begin{equation}
  \hat{y} \;=\; \arg\max_{c \in \mathcal{Y}}\ \cos\!\big(u,\, p_c\big).
  \label{eq:classify}
\end{equation}
The prototype $p_c$ is the embedding-space projection of a stored raw prototype
$\mu_c \in \mathbb{R}^{D}$, whose construction and update we defer to
\S\ref{sec:drift}.

\subsubsection{Context-to-Target Prediction}
\label{sec:prediction}

The core of our framework is the relation that predicts the target from the
context in the latent space. A predictor $q_\phi(\cdot)$, with trainable
parameters $\phi$, estimates the target from the context representation:
\begin{equation}
  \hat{u} \;=\; q_\phi(v).
  \label{eq:predict}
\end{equation}
The prediction is aligned to a teacher target $\tilde{u}$. The teacher is an
exponential moving average (EMA) copy $g_{\xi}(\cdot)$ of the encoder, whose
parameters are updated at each training step as $\xi \leftarrow m\,\xi +
(1-m)\,\theta$ with EMA coefficient $m \in [0,1)$, and no gradient flows along this
path; using an EMA copy rather than the online encoder gives a slowly moving,
stable target while $g_\theta$ is still being learned. Using the label, we mix the
normalized teacher output with the corresponding class prototype to stabilize the
semantic target:
\begin{equation}
  \tilde{u} \;=\; \mathrm{normalize}\!\Big(
    \lambda\, \bar{g}_{\xi}(u^{\mathrm{raw}}) \;+\; (1-\lambda)\, p_{y}
  \Big),
  \qquad
  \bar{g}_{\xi}(z) = \mathrm{normalize}\!\big(g_{\xi}(z)\big),
  \label{eq:teacher-target}
\end{equation}
\begin{equation}
    \mathcal{L}_{\mathrm{pred}} \;=\; 1 - \cos\!\big(\hat{u},\ \mathrm{sg}(\tilde{u})\big),
    \label{eq:pred-loss}
\end{equation}
where $y$ is the input label, $p_y$ is the prototype of class $y$, $\lambda \in
[0,1]$ is the mixing ratio between the normalized teacher output
$\bar{g}_{\xi}(u^{\mathrm{raw}})$ and the class prototype,
$\mathrm{normalize}(\cdot)$ denotes $\ell_2$ normalization, and
$\mathrm{sg}(\cdot)$ is the stop-gradient operator. This loss aligns $g_\theta$ and
$q_\phi$ so that the target indicated by the context is consistently reconstructed,
capturing the semantic structure that should persist across domains. This same
context-to-target predictive relation is later reused in \S\ref{sec:drift} to
measure the drift $\delta_{s,c}^{(t)}$ by re-inputting stored past context
summaries.

\subsection{Drift-Adaptive Consolidation}
\label{sec:drift}
This section concretely constructs the anchor $a_c^{(t-1)}$ and the drift
$\delta_{s,c}^{(t)}$ introduced conceptually in \S\ref{sec:problem-setup}, and
reuses the predictive relation of \S\ref{sec:evidence-prediction} to measure the
drift of past concepts during training. The measured drift is then used in two
ways: the consolidation loss concentrates preservation on high-drift pairs, while
prototype construction lowers the contribution of high-drift domains.
Notationally, we distinguish the raw token space produced by the backbone,
$\mathbb{R}^{D}$, from the embedding space to which the encoder maps,
$\mathbb{R}^{d}$. Prototypes are stored as class means in the raw space and are
mapped to the embedding space by the encoder at the time of use.

\subsubsection{Anchor Construction}
\label{sec:anchor}
At the end of training on each domain $s$, we store for every observed class $c$
its raw target class mean $m_{s,c} \in \mathbb{R}^{D}$, the mean of the raw
target representations $u^{\mathrm{raw}}$ over the samples of class $c$ in domain
$s$. The prototype of class $c$ is maintained as a raw class mean
$\mu_c \in \mathbb{R}^{D}$ that combines these per-domain means over the domains
in which the class was observed. The embedding prototype used for classification is its
mapping through the current encoder, $p_c = g_\theta(\mu_c)$. Immediately before
training on domain $t$ (the domain boundary), we take a fixed snapshot of the
teacher encoder $g_{\xi^{(t-1)}}$ and the raw prototype $\mu_c^{(t-1)}$ at that
point, and define the anchor of class $c$ as the normalized embedding
representation that this teacher produces for the raw prototype:
\begin{equation}
  a_c^{(t-1)}
  \;=\;
  \mathrm{normalize}\!\Big(g_{\xi^{(t-1)}}\!\big(\mu_c^{(t-1)}\big)\Big)
  \;\in\; \mathbb{R}^{d},
  \qquad
  \mu_c^{(t-1)} \in \mathbb{R}^{D},
  \label{eq:anchor-def}
\end{equation}
where $\xi^{(t-1)}$ denotes the teacher parameters fixed immediately before
training on domain $t$. The anchor is fixed once at the domain boundary and does
not change during training on domain $t$; this is the concrete realization of the
``reference that does not move'' described in \S\ref{sec:problem-setup}. Because
the anchor is defined at the class level, the per-domain difference in drift
arises not from the anchor but from the past representation given as input.

\subsubsection{Drift Measurement via Replay}
\label{sec:drift-measure}
Measuring drift requires re-evaluating a past concept under the current model
without its original images. To this end, at the end of training on each domain we
store, for every observed class, a raw context summary $\nu_{s,c} \in
\mathbb{R}^{D}$, the mean of the raw context representations $v^{\mathrm{raw}}$ over
the samples of class $c$ in domain $s$. This is stored alongside the raw target class mean $m_{s,c}$;
the two summaries are the only state we retain for each domain--class pair.
At the current step $t$, we reconstruct the target of a past concept by passing
its stored context summary through the current encoder and predictor,
\begin{equation}
  r_{s,c}^{(t)} \;=\; q_\phi\!\big(g_\theta(\nu_{s,c})\big) \;\in\; \mathbb{R}^{d},
  \label{eq:replay}
\end{equation}
which applies the same context-to-target relation \eqref{eq:predict} used during
current-domain training, now to the past summary rather than to a fresh image. The
drift of class $c$ from domain $s$ is then the cosine distance between this
reconstruction and the fixed anchor,
\begin{equation}
  \delta_{s,c}^{(t)} \;=\; 1 - \cos\!\big(r_{s,c}^{(t)},\ a_c^{(t-1)}\big)
  \ \ \text{for } s < t,
  \qquad
  \delta_{t,c}^{(t)} \;\coloneqq\; 0,
  \label{eq:drift}
\end{equation}
which is nonnegative and serves as a surrogate for whether the current model still
maps the past context to a representation aligned with the class anchor; the
current domain carries no accumulated drift yet, so its drift is defined as zero.
Because the same predictive relation both learns representations on the current
domain in \eqref{eq:pred-loss} and serves as the meter of drift for past
summaries, measurement and learning act through one function.

\subsubsection{Adaptive Consolidation}
\label{sec:adaptive-cons}

The consolidation loss constrains $g_\theta$ and $q_\phi$ so that the past
context-to-target relation is preserved under the current model, that is, so that
the measured drift is reduced. Let $\mathcal{S}^{(t)} = \{(s,c) : s < t,\
\nu_{s,c}\ \text{stored}\}$ be the set of past domain--class pairs whose summaries
are currently stored. Rather than treating all pairs equally, we place greater
preservation pressure on pairs that have drifted more,
\begin{equation}
  \omega_{s,c}
  \;=\;
  \frac{\big(\mathrm{sg}(\delta_{s,c}^{(t)}) + \epsilon\big)^{\gamma}}
  {\frac{1}{|\mathcal{S}^{(t)}|}
   \sum_{(s',c') \in \mathcal{S}^{(t)}}
   \big(\mathrm{sg}(\delta_{s',c'}^{(t)}) + \epsilon\big)^{\gamma}},
  \label{eq:drift-weight}
\end{equation}
\begin{equation}
  \mathcal{L}_{\mathrm{cons}}
  \;=\;
  \frac{1}{|\mathcal{S}^{(t)}|}
  \sum_{(s,c) \in \mathcal{S}^{(t)}}
  \omega_{s,c}\; \delta_{s,c}^{(t)},
  \label{eq:cons-loss}
\end{equation}
where $\gamma \ge 0$ controls how sharply the weighting responds to drift,
$\mathrm{sg}(\cdot)$ is the stop-gradient operator, and $\epsilon > 0$ is a small
constant for numerical stability, especially when some past pairs have near-zero
drift.

The weight $\omega_{s,c}$ has two properties. First, the drift is detached through
$\mathrm{sg}$, so gradients flow only through the $\delta_{s,c}^{(t)}$ factor in
\eqref{eq:cons-loss}: the weight rescales each pair by its drift magnitude but is
not itself a learning signal. Second, $\omega_{s,c}$ is normalized to unit mean
over $\mathcal{S}^{(t)}$, so the overall loss scale is preserved as $\gamma$
varies, and dividing the sum by $|\mathcal{S}^{(t)}|$ keeps it stable as past pairs
accumulate. At $\gamma = 0$ the weights reduce to $\omega_{s,c} = 1$ and all pairs
are preserved uniformly; as $\gamma$ grows, preservation concentrates on the
highest-drift pairs.

\subsubsection{Robust Prototype Construction}
\label{sec:robust-proto}

The same drift signal also governs how the prototype is built. Recall from the Anchor Construction above that the raw prototype $\mu_c^{(t)}$
combines the per-domain target means $m_{s,c} \in \mathbb{R}^{D}$; we now specify that combination. When a
domain has drifted far, its target mean, once projected through the current
encoder, is likely misaligned with the stable semantics of that class, so
combining it with equal weight would let it contaminate the prototype. We
therefore down-weight each domain by its measured drift,
\begin{equation}
  \psi_{s,c}
  \;=\;
  \mathrm{clip}\!\left(
    \left(
      \frac{\mathrm{sg}(\delta_{s,c}^{(t)}) + \epsilon}
           {\bar{\delta}_{c}^{(t)} + \epsilon}
    \right)^{-\rho},
    \ \tfrac{1}{\kappa},\ \kappa
  \right),
  \qquad
  \bar{\delta}_{c}^{(t)}
  \;=\;
  \frac{1}{|\mathcal{H}_c^{(t)}|}
  \sum_{s \in \mathcal{H}_c^{(t)}} \mathrm{sg}\!\big(\delta_{s,c}^{(t)}\big),
  \label{eq:proto-weight}
\end{equation}
\begin{equation}
  \mu_c^{(t)}
  \;=\;
  \sum_{s \in \mathcal{H}_c^{(t)}}
  \pi_{s,c}\; m_{s,c},
  \qquad
  \pi_{s,c}
  \;=\;
  \frac{\psi_{s,c}}{\sum_{s' \in \mathcal{H}_c^{(t)}} \psi_{s',c}},
  \label{eq:proto-build}
\end{equation}
where $\mathcal{H}_c^{(t)} = \{s \le t : m_{s,c}\ \text{stored}\}$ is the set of
domains in which class $c$ has been observed, $\bar{\delta}_{c}^{(t)}$ is the mean
drift over those domains, $\rho \ge 0$ controls how strongly reliability responds
to drift, and $\kappa \ge 1$ bounds each unnormalized weight $\psi_{s,c}$ to
$[\,1/\kappa,\ \kappa\,]$. As in \eqref{eq:drift-weight}, the drift is detached
through $\mathrm{sg}$, so $\psi_{s,c}$ reweights without acting as a learning
signal, and the weights are normalized to sum to one per class. Unlike the consolidation weight, $\mathcal{H}_c^{(t)}$ includes the current domain,
whose drift is zero by construction ($\delta_{t,c}^{(t)} = 0$). This turns
$\bar{\delta}_{c}^{(t)}$ into a per-class reference against the latest observation:
ignoring the small stabilizer $\epsilon$, a domain that has drifted more than the
class average has a ratio $\delta_{s,c}^{(t)} / \bar{\delta}_{c}^{(t)} > 1$ and,
under the negative exponent $-\rho$, contributes less, while stable domains receive
larger prototype weights. At $\rho = 0$ all weights equal one and $\mu_c^{(t)}$
reduces to the uniform mean of its per-domain means, mirroring the $\gamma = 0$
case of consolidation.

A single drift signal $\delta_{s,c}^{(t)}$ thus plays two complementary roles.
Consolidation \eqref{eq:cons-loss} places stronger preservation pressure on
high-drift pairs, so that concepts at higher risk of forgetting receive stronger
protection; prototype construction \eqref{eq:proto-build} trusts high-drift domains
less, so that the class reference stays stable. The former decides where to
concentrate preservation, the latter which representation to trust: consolidation
reduces drift over training, while robust construction keeps the prototype free of
the contamination that residual drift would otherwise introduce.

\subsection{Training Objective}
\label{sec:training}

At each training step, the model minimizes a weighted sum of three losses,
\begin{equation}
  \mathcal{L}
  \;=\;
  w_{\mathrm{cls}}\,\mathcal{L}_{\mathrm{cls}}
  \;+\;
  w_{\mathrm{pred}}\,\mathcal{L}_{\mathrm{pred}}
  \;+\;
  w_{\mathrm{cons}}\,\mathcal{L}_{\mathrm{cons}},
  \label{eq:total-loss}
\end{equation}
where $w_{\mathrm{cls}}, w_{\mathrm{pred}}, w_{\mathrm{cons}} \ge 0$ weight the
respective terms. The classification loss is the cross-entropy over prototype
logits, obtained by scaling the cosine similarity between the target embedding $u$
and each class prototype $p_c$ by a learnable logit scale $\tau$,
\begin{equation}
  \mathcal{L}_{\mathrm{cls}}
  \;=\;
  -\log
  \frac{\exp\!\big(\tau\,\cos(u, p_{y})\big)}
       {\sum_{c \in \mathcal{Y}} \exp\!\big(\tau\,\cos(u, p_{c})\big)},
  \label{eq:cls-loss}
\end{equation}
where $y$ is the ground-truth label, $p_c$ is the class prototype defined in
\S\ref{sec:anchor}, and $\tau > 0$ is a learnable logit scale. The prediction loss
$\mathcal{L}_{\mathrm{pred}}$ and the consolidation loss
$\mathcal{L}_{\mathrm{cons}}$ are as defined in \eqref{eq:pred-loss} and
\eqref{eq:cons-loss}, respectively.

The three terms play distinct roles. The classification loss preserves
discriminability on the current domain, the prediction loss learns the
context-to-target relation, and the consolidation loss keeps that relation from
breaking down on past domain--class pairs. On the first domain, no past summary has
been stored, so $\mathcal{S}^{(1)} = \varnothing$ and $\mathcal{L}_{\mathrm{cons}}
= 0$; consolidation becomes active from the second domain onward.

Training follows the domain sequence, with the pretrained backbone kept frozen
throughout. At each domain boundary ($t \ge 2$), we fix the anchor $a_c^{(t-1)}$
from the snapshot teacher as described in \S\ref{sec:anchor}. During training on
domain $t$, we update the online encoder $g_\theta$, the predictor $q_\phi$, and
the logit scale $\tau$ by minimizing \eqref{eq:total-loss}, while the teacher
parameters $\xi$ are updated by EMA after each step. When training on domain $t$
ends, we store the raw target mean $m_{t,c}$ and raw context summary $\nu_{t,c}$
for each observed class, and recompute the raw prototypes $\mu_c^{(t)}$ via
\eqref{eq:proto-build}, with the drift $\delta_{s,c}^{(t)}$ evaluated at the end
of training on domain $t$, for use in subsequent domains.

\section{Experiments}
\label{sec:experiments}

\newcommand{\std}[1]{{\scriptsize$\pm$#1}}
\begin{table}[t]
  \caption{Experiment results. For each benchmark we report the AvgAcc ($\uparrow$) and AvgF ($\downarrow$), in percent.
  Bold marks the best and underline the second best in each column.}
  \label{tab:headings}
  \centering
  \setlength{\tabcolsep}{3pt}
  \resizebox{\textwidth}{!}{%
  \begin{tabular}{lcccccccc}
    \toprule
    \multirow{2}{*}{Model}
      & \multicolumn{2}{c}{DomainNet} & \multicolumn{2}{c}{CORe50}
      & \multicolumn{2}{c}{OfficeHome} & \multicolumn{2}{c}{CDDB-Hard}\\
    \cmidrule(lr){2-3} \cmidrule(lr){4-5} \cmidrule(lr){6-7} \cmidrule(lr){8-9}
      & AvgAcc\,$\uparrow$ & AvgF\,$\downarrow$ & AvgAcc\,$\uparrow$ & AvgF\,$\downarrow$
      & AvgAcc\,$\uparrow$ & AvgF\,$\downarrow$ & AvgAcc\,$\uparrow$ & AvgF\,$\downarrow$\\
    \midrule
    DIDD \cite{oh2026asymmetric}        & 9.42\std{3.23} & 4.75\std{2.87} & 55.9\std{2.07} & 22.6\std{1.85} & 29.5\std{2.60} & 3.9\std{1.60} & 54.5\std{1.54} & \textbf{1.7}\std{0.87}\\
    CODA-Prompt \cite{smith2023coda} & 53.6\std{2.33} & 21.5\std{3.27} & 99.1\std{0.15} & 0.8\std{0.16} & 87.2\std{0.13} & 4.7\std{0.26} & 67.8\std{1.01} & 12.9\std{0.92}\\
    OS-Prompt \cite{kim2024one}   & 52.1\std{1.59} & 23.3\std{1.13} & 99.2\std{0.08} & 0.8\std{0.08} & 86.6\std{0.17} & 5.5\std{0.14} & 66.9\std{0.73} & 20.2\std{0.51}\\
    OS-Prompt++ \cite{kim2024one} & 51.4\std{1.18} & 24.3\std{1.87} & 99.2\std{0.03} & 0.8\std{0.02} & 86.3\std{0.47} & 5.2\std{0.46} & 67.5\std{1.76} & 17.9\std{2.14}\\
    MoE \cite{yu2024boosting}        & \underline{63.6}\std{0.82} & 17.9\std{2.23} & 99.3\std{0.02} & 0.5\std{0.08} & \underline{87.5}\std{0.05} & 1.8\std{0.27} & \textbf{81.4}\std{0.48} & 8.3\std{1.53}\\
    GraphKeeper \cite{guo2026graphkeeper} & 51.8\std{0.51} & 8.6\std{0.47} & 98.2\std{0.07} & 0.3\std{0.04} & 22.6\std{0.22} & 0.6\std{0.13} & 61.6\std{0.86} & \underline{1.9}\std{0.92}\\
    SOYO \cite{wang2025boosting}              & 51.2\std{0.93} & \textbf{2.0}\std{0.55} & 99.6\std{0.04} & \textbf{0.1}\std{0.03} & 72.3\std{0.08} & \underline{0.3}\std{0.20} & 78.3\std{1.54} & 2.4\std{0.73}\\
    DUCT \cite{zhou2025dual}              & \textbf{66.4}\std{1.02} & 4.1\std{1.13} & 94.4\std{0.05} & 0.4\std{0.08} & 87.2\std{0.10} & 4.3\std{0.62} & 71.2\std{3.58} & 39.0\std{4.88}\\
    \midrule
    \textbf{Ours}               & 61.8\std{0.97} & \underline{4.0}\std{1.07} & \textbf{99.7}\std{0.04} & \underline{0.3}\std{0.03} & \textbf{87.8}\std{0.04} & \textbf{0.2}\std{0.50} & \underline{79.4}\std{0.33} & 6.7\std{0.87}\\
    \bottomrule
  \end{tabular}}
\end{table}

\subsection{Experiment Results}
\label{sec:main-results}

We evaluate our method on four standard DIL benchmarks whose domains are presented
sequentially: DomainNet \cite{peng2019moment}, CORe50 \cite{lomonaco2017core50},
Office-Home \cite{venkateswara2017deep}, and CDDB-Hard \cite{li2023continual}. We
report the two standard DIL metrics: the final average accuracy AvgAcc, averaged
over all seen domains after the last stage, and the forgetting AvgF, the average
drop between the peak accuracy each domain reached and its accuracy at the final
stage. Their formal definitions, together with dataset details, data splits, and
the hyperparameter settings, are provided in
Appendices~\ref{app:metrics}--\ref{app:implementation}. All results are
averaged over three runs with different random seeds.
\begin{table}[t]
  \caption{Peak GPU memory usage on DomainNet (GB).}
  \label{tab:vram}
  \centering
  \setlength{\tabcolsep}{12pt}
  \resizebox{\textwidth}{!}{%
  \begin{tabular}{lccccccccc}
    \toprule
    Model & DIDD & CODA-Prompt & OS-Prompt & OS-Prompt++ & MoE & GraphKeeper & SOYO & DUCT & \textbf{Ours} \\
    \midrule
    VRAM (GB) & 6.38 & 19.99 & 19.79 & 20.21 & 32.00 & $\mathbf{2.00}$ & 15.23 & 14.07 & \underline{3.00} \\
    \bottomrule
  \end{tabular}}%
\end{table}
Table~\ref{tab:headings} reports AvgAcc and AvgF across the four benchmarks. Our
method achieves the highest accuracy on CORe50 and Office-Home and remains
competitive on DomainNet and CDDB-Hard, while keeping forgetting low on every
benchmark. In contrast, the accuracy-leading baselines show pronounced weaknesses
on other benchmarks: DUCT achieves the best accuracy on DomainNet but reaches
$39.0\%$ forgetting on CDDB-Hard, and MoE, the strongest method on CDDB-Hard,
reaches $17.9\%$ forgetting on DomainNet. These figures indicate substantial
degradation on previously learned domains as new ones are learned. Our method
instead maintains a strong accuracy--forgetting balance across all four
benchmarks. Its lowest per-benchmark accuracy, $61.8\%$ on DomainNet, trails the
accuracy leaders DUCT and MoE but remains higher than every other compared method,
and it comes with only $4.0\%$ forgetting---an order of magnitude below DUCT's
worst-case $39.0\%$. High accuracy and low forgetting are thus achieved together,
and consistently, rather than by improving one benchmark at the expense of severe
forgetting on another.

\begin{figure}[t]
\centering
\includegraphics[width=0.5\textwidth]{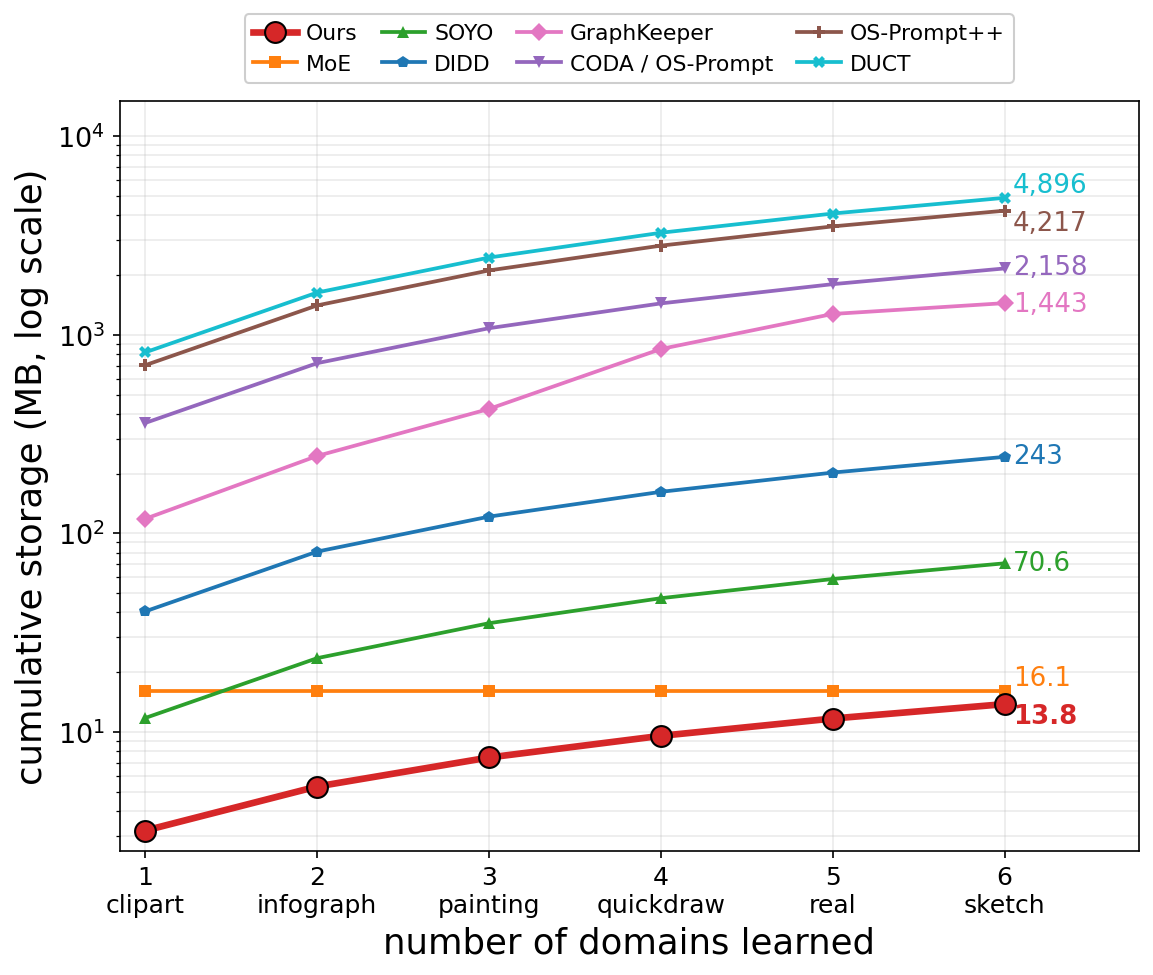}
\caption{Cumulative storage (MB, log scale) on DomainNet as the number of
learned domains grows. Numbers at the right denote the final storage after all
six domains.} \label{fig:memory}
\end{figure}

\subsection{Efficiency Analysis}
\label{sec:efficiency}

We analyze efficiency along two axes: the GPU memory used during training and the
storage that accumulates as domains grow. Table~\ref{tab:vram} reports
training-time GPU memory on DomainNet, measured under the same protocol, hardware,
batch size, and resolution. Our method uses only $3.0$~GB, $4.7\times$ and
$10.7\times$ less than its strongest accuracy competitors DUCT ($14.07$~GB) and
MoE ($32.00$~GB). This follows from our design: the backbone stays frozen with no
gradients through it, and only a lightweight encoder and predictor are trained.
GraphKeeper uses less memory and DIDD remains moderate, but both are markedly
weaker in accuracy on several benchmarks.

Figure~\ref{fig:memory} shows, on a log scale, the storage each method accumulates
as domains grow. Because we store neither raw images nor a generative model, but
only a compact context summary and target mean per domain--class pair, our storage
grows only mildly, reaching $13.78$~MB after the sixth domain---the smallest among
all compared methods, and roughly $355\times$ smaller than the heaviest, DUCT.
Prompt- and feature-based methods accumulate far more (up to $4{,}217$~MB), and
the gap widens with the number of domains, making our advantage more pronounced in
long-term continual learning. Overall, our method remains competitive with the
strongest accuracy baselines while using $4.7\times$--$10.7\times$ less training
memory and the smallest storage among all compared methods, occupying a favorable
position in the accuracy--efficiency trade-off.

\begin{table}[t]
\centering
\caption{Ablation study. Each variant removes one
component of the full model. AvgAcc and AvgF are the final average accuracy and
forgetting (\%).}
\label{tab:ablation}
\begin{tabular}{l cc cc}
\toprule
& \multicolumn{2}{c}{DomainNet} & \multicolumn{2}{c}{CDDB-Hard} \\
\cmidrule(lr){2-3} \cmidrule(lr){4-5}
Variant & AvgAcc $\uparrow$ & AvgF $\downarrow$ & AvgAcc $\uparrow$ & AvgF $\downarrow$ \\
\midrule
\textbf{Full (Ours)}                         & \textbf{61.80} & \textbf{4.12}  & \textbf{79.40} & \textbf{6.70} \\
\midrule
w/o $\mathcal{L}_{\mathrm{cons}}$   & 58.18 & 14.50 & 76.82 & 7.74 \\
w/o $\mathcal{L}_{\mathrm{pred}}$   & 59.83 & 5.53 & 76.28 & 8.22\\
w/o adaptive cons.\ ($\gamma{=}0$)  & 58.25 & 8.56  & 78.46 & 6.88 \\
w/o robust proto.\ ($\rho{=}0$)     & 58.80 & 7.07  & 77.71 & 7.28 \\
\bottomrule
\end{tabular}
\end{table}

\subsection{Ablation Study}
\label{sec:ablation}
We analyze the contribution of each component on DomainNet and CDDB-Hard.
Table~\ref{tab:ablation} reports the final AvgAcc and AvgF when the prediction
loss, the consolidation loss, the drift-adaptive weighting, and the robust
prototype weighting are each removed from the full model. The three drift-driven
components affect forgetting far more than accuracy. On DomainNet, removing the
consolidation loss is the largest effect, surging forgetting from $4.12\%$ to
$14.50\%$ (roughly $3.5\times$), confirming it as the core mechanism for
preserving prior knowledge. Treating all past domain--class pairs uniformly
($\gamma = 0$) and disabling the robust prototype weighting ($\rho = 0$) increase
forgetting by about $2.1\times$ and $1.7\times$, respectively, indicating that both drift-aware weighting mechanisms contribute to reducing forgetting.
Removing the prediction loss instead costs accuracy at least as much as
stability, consistent with $\mathcal{L}_{\mathrm{pred}}$ learning the
context-to-target relation on which both the representation and the drift
measurement rest. The same trends hold on CDDB-Hard, with smaller absolute
changes given its fewer domains.

\begin{figure}[b]
\centering
\includegraphics[width=0.5\textwidth]{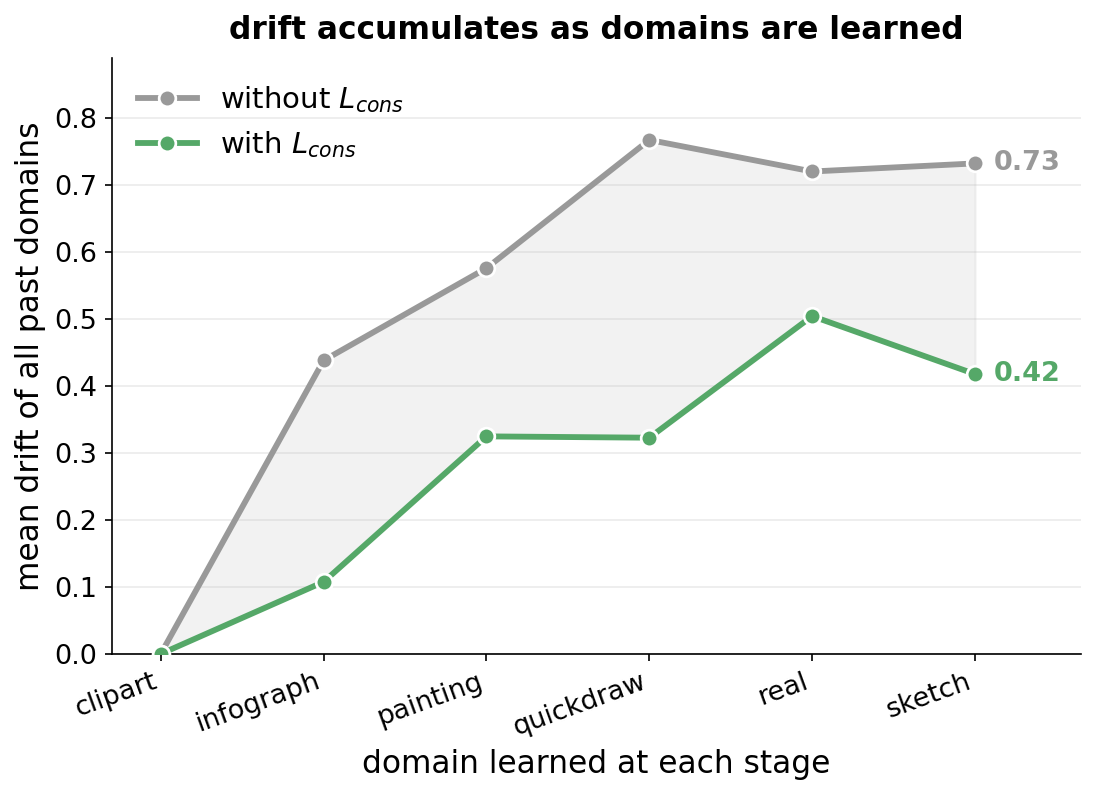}
\caption{Mean drift over all past domains at each training stage on DomainNet,
with and without the consolidation loss $\mathcal{L}_{\mathrm{cons}}$.} \label{fig:anchor_gamma}
\end{figure}

\begin{figure}[t]
\centering
\includegraphics[width=0.7\textwidth]{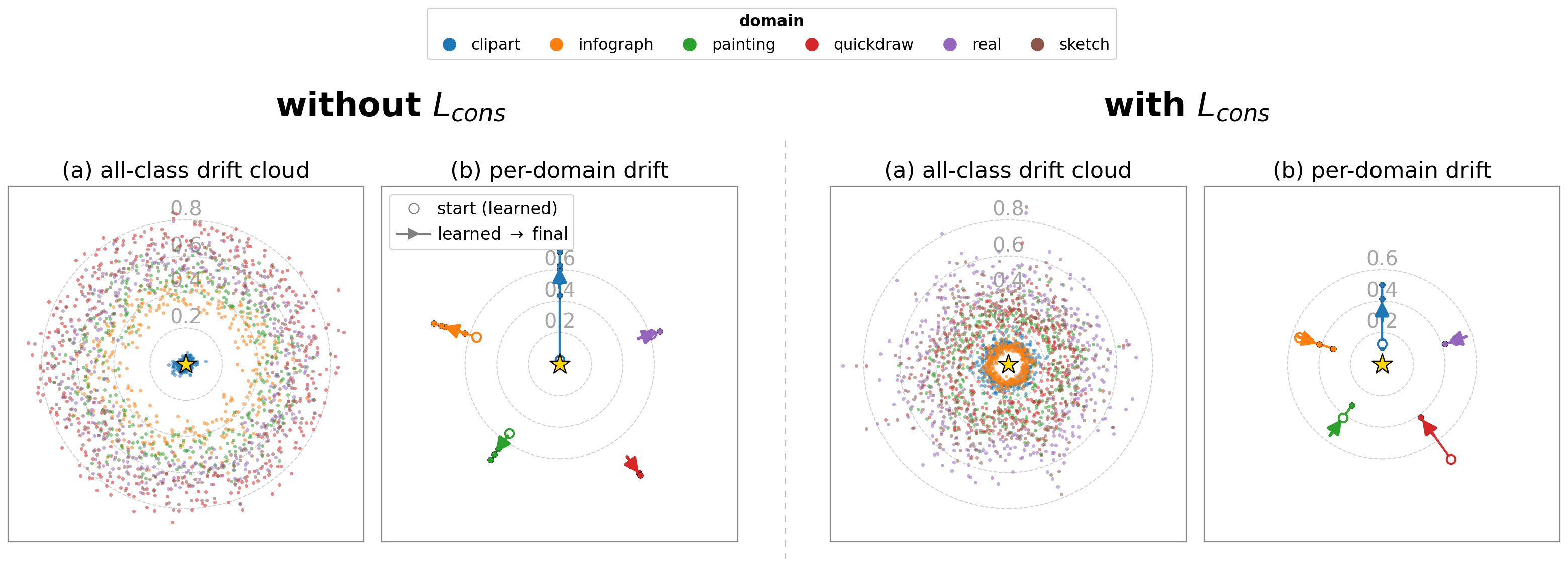}
\caption{Polar visualization of drift with and without
$\mathcal{L}_{\mathrm{cons}}$, where all class anchors are placed at the center
({\color{yellow!80!orange}$\bigstar$}) and the radius denotes the drift magnitude.}
\label{fig:anchor_combined}
\end{figure}

While Table~\ref{tab:ablation} quantifies these effects, Figures~\ref{fig:anchor_gamma}
and~\ref{fig:anchor_combined} show them directly in the representation space. We
measure the drift of a class as the cosine distance between the target
reconstructed from its stored context and its anchor; for visualization, we plot
every class by its distance to its own anchor and place all anchors at a single
center, so that the radius denotes the magnitude of drift.
Figure~\ref{fig:anchor_gamma} plots the average drift over past domains at each
training stage. Without the consolidation loss, past representations move steadily
away from their anchors as domains accumulate, reaching a final average drift of
$0.73$, whereas the consolidation loss suppresses it to around $0.42$. The suppression is largest at the Quickdraw stage, where the measured drift also peaks, showing that the consolidation loss has its strongest effect when representation change is substantial.
Figure~\ref{fig:anchor_combined} shows the same phenomenon through a polar
visualization centered at each class anchor. In (a), each point is the drift of
one class; without the consolidation loss the points of later domains spread
widely into the outer region, whereas the consolidation loss shifts most points
toward the center. In (b), each spoke is one past domain, and the trajectory along
it depicts the change in drift from immediately after learning that domain to the
final state. Without the consolidation loss, all trajectories extend outward as
subsequent domains are learned; with it, the trajectories are shorter overall, and
those of high-drift domains point back inward. Thus, the consolidation loss suppresses further drift while pulling already deviated representations back toward their anchors.

\section{Conclusion}
\label{sec:conclusion}

This work addresses nonuniform forgetting across domains and classes in
domain-incremental learning by using representation drift as a measurable
surrogate for forgetting risk. With a frozen backbone and a JEPA-based
predictive relation, we measure how far each past domain--class concept deviates
from its anchor during training without storing raw data. The measured drift
guides two complementary mechanisms: adaptive consolidation emphasizes
high-drift pairs, while robust prototype construction reduces their influence
on class references.
Across four benchmarks, our method maintains a strong accuracy--forgetting
balance while remaining resource efficient. It achieves the best accuracy on
two benchmarks and competitive performance on the others, with consistently low
forgetting. By training only a lightweight encoder and predictor and storing
compact domain--class summaries, the method requires low training memory and
small storage overhead.

\section*{Acknowledgments}
This work was supported partly by the BK21 FOUR Project and partly by the Technology development Program RS-2025-25454734 funded by the Ministry of SMEs and Startups (MSS, Korea).

\bibliographystyle{splncs04}
\bibliography{main}

\clearpage
\appendix
\begin{center}
  {\Large\bfseries Supplementary Material}
\end{center}

\section{Evaluation Metrics}
\label{app:metrics}
We give the formal definitions of the two metrics reported in the main paper.
Let $a_{s,j}$ denote the accuracy of the model trained up to stage $s$,
evaluated on domain $j$. At the final stage $T$,
\begin{equation}
  A_T = \frac{1}{T}\sum_{j=1}^{T} a_{T,j},
  \qquad
  F_T = \frac{1}{T-1}\sum_{j=1}^{T-1}
  \Big(\max_{s\in\{j,\dots,T-1\}} a_{s,j} - a_{T,j}\Big).
  \label{eq:metrics}
\end{equation}
$A_T$ (AvgAcc) is the empirical estimate, at the final stage, of the
seen-domain average objective defined in Eq.~\eqref{eq:dil-objective}, and $F_T$
(AvgF) measures forgetting as the average drop between the peak accuracy each
domain reached and its final accuracy.

\section{Datasets and Splits}
\label{app:datasets}
\begin{table}[h]
  \caption{Benchmark statistics. Domains are presented sequentially in the
  listed order.}
  \label{tab:datasets}
  \centering
  \footnotesize
  \setlength{\tabcolsep}{4pt}
  \begin{tabular}{lccp{5.6cm}}
    \toprule
    Benchmark & \#Domains & \#Classes & Domain order \\
    \midrule
    DomainNet   & 6  & 345 & clipart, infograph, painting, quickdraw, real, sketch \\
    CORe50      & 11 & 50  & s1, s2, s3, s4, s5, s6, s7, s8, s9, s10, s11 \\
    Office-Home & 4  & 65  & Art, Clipart, Product, Real World \\
    CDDB-Hard   & 5  & 2   & gaugan, biggan, wild, whichfaceisreal, san \\
    \bottomrule
  \end{tabular}
\end{table}
\paragraph{DomainNet~\cite{peng2019moment}} contains six visually distinct
domains over 345 shared classes. All images are re-split per domain and class
into train/validation/test with an approximate 8:1:1 ratio, stratified by
class.
\paragraph{CORe50~\cite{lomonaco2017core50}} consists of 50 object classes
recorded in eleven sessions with varying backgrounds and conditions; each
session is treated as one domain. Images are re-split per domain and class into
train/validation/test at approximately 8:1:1.
\paragraph{Office-Home~\cite{venkateswara2017deep}} covers 65 classes across
four domains. Since the dataset ships no official split, images are re-split
per domain and class into train/validation/test at approximately 8:1:1.
\paragraph{CDDB-Hard~\cite{li2023continual}} is the hard track of the continual
deepfake detection benchmark: binary real/fake classification over five
generator domains, following the standard sequence used by prior DIL work. We
use the official per-domain train/val folders, with val serving as the test
set, as is standard practice on this benchmark.

All images are resized to $224 \times 224$ and normalized with ImageNet
statistics. The validation split, where present, is used only for early
stopping (Sec.~\ref{app:implementation}) and never for training.

\section{Implementation Details}
\label{app:implementation}
\paragraph{Architecture.} All experiments use a frozen DINOv2
ViT-B/14~\cite{oquab2023dinov2} backbone ($D = 768$); no gradient flows through
it. The online encoder $g_\theta$ is a two-layer residual MLP and the predictor
$q_\phi$ is a two-layer MLP, both operating at the backbone feature dimension
($d = D = 768$). The teacher $g_\xi$ is an EMA copy of $g_\theta$ with momentum
$m = 0.996$. Target/context pooling uses the soft attention weighting of
Eq.~\eqref{eq:pool}.
\paragraph{Optimization.} We train with AdamW (learning rate $10^{-3}$, weight
decay $10^{-4}$), batch size 128, a cosine learning-rate schedule without
warmup, gradient clipping at max-norm $1.0$, and mixed precision. Early
stopping monitors validation accuracy with patience 5. The classification and
prediction loss weights are $w_{\mathrm{cls}} = w_{\mathrm{pred}} = 1$
throughout; the per-benchmark consolidation weight $w_{\mathrm{cons}}$,
drift-weighting sharpness $\gamma$, and number of epochs are listed in
Table~\ref{tab:hyper}.
\paragraph{Method hyperparameters.} The teacher-mixing ratio $\lambda$ is $0.5$; the logit scale $\tau$ is learnable with
initial value $10.0$; the reliability exponent is
$\rho = 2$ with clip bound $\kappa = 10$; the numerical stabilizer is
$\epsilon = 10^{-6}$.
\begin{table}[h]
  \caption{Per-benchmark training settings. All other hyperparameters are
  shared across benchmarks.}
  \label{tab:hyper}
  \centering
  \footnotesize
  \begin{tabular}{lcccc}
    \toprule
    & DomainNet & CORe50 & Office-Home & CDDB-Hard \\
    \midrule
    Epochs per domain      & 10  & 10 & 30  & 20 \\
    $w_{\mathrm{cons}}$    & 300 & 3  & 300 & 30 \\
    $\gamma$               & 2   & 1  & 1   & 2  \\
    \bottomrule
  \end{tabular}
\end{table}

$w_{\mathrm{cons}}$ differs across benchmarks because the scale of the
consolidation term itself does: benchmarks with strong distributional
heterogeneity (e.g., DomainNet) exhibit large drift, so a larger weight is
needed for the preservation pressure to remain comparable to the current-domain
losses, whereas benchmarks whose domains are visually close (e.g., CORe50),
even with many stages, accumulate little drift and require only a small weight.
Values were selected by a grid over $w_{\mathrm{cons}} \in \{10, 30, 100,
300\}$ and $\gamma \in \{0, 1, 2\}$ on seed~0
(Sec.~\ref{app:sensitivity}). All experiments run on a single NVIDIA A100 GPU,
and all results are averaged over three runs with different random seeds.

\section{Sensitivity to the Consolidation Weight}
\label{app:sensitivity}
\begin{table}[t]
  \centering
  \caption{Sensitivity to the consolidation weight $w_{\mathrm{cons}}$ on
  DomainNet. Each cell reports AvgAcc\,/\,AvgF in percent; \textbf{bold} marks
  the adopted configuration ($w_{\mathrm{cons}}{=}300$, $\gamma{=}2$).}
  \label{tab:sens-wcons-domainnet}
  \setlength{\tabcolsep}{5pt}
  \begin{tabular}{lcccc}
    \toprule
    $\gamma \,\backslash\, w_{\mathrm{cons}}$ & $10$ & $30$ & $100$ & $300$ \\
    \midrule
    $\gamma{=}0$ & 60.59\,/\,11.52 & 60.63\,/\,11.23 & 61.47\,/\,9.87  & 58.25\,/\,8.56 \\
    $\gamma{=}1$ & 59.80\,/\,12.78 & 60.51\,/\,11.63 & 61.06\,/\,10.52 & 60.51\,/\,6.40 \\
    $\gamma{=}2$ & 59.90\,/\,12.66 & 60.53\,/\,11.61 & 61.19\,/\,9.85  & \textbf{61.80\,/\,4.12} \\
    \bottomrule
  \end{tabular}
\end{table}
\begin{table}[t]
  \centering
  \caption{Sensitivity to the consolidation weight $w_{\mathrm{cons}}$ on
  Office-Home. Each cell reports AvgAcc\,/\,AvgF in percent; \textbf{bold}
  marks the adopted configuration ($w_{\mathrm{cons}}{=}300$, $\gamma{=}1$).}
  \label{tab:sens-wcons-officehome}
  \setlength{\tabcolsep}{5pt}
  \begin{tabular}{lcccc}
    \toprule
    $\gamma \,\backslash\, w_{\mathrm{cons}}$ & $10$ & $30$ & $100$ & $300$ \\
    \midrule
    $\gamma{=}0$ & 87.12\,/\,1.82 & 86.83\,/\,2.14 & 87.75\,/\,0.21 & 87.68\,/\,0.50 \\
    $\gamma{=}1$ & 84.93\,/\,4.09 & 87.16\,/\,0.99 & 87.51\,/\,0.80 & \textbf{87.80\,/\,0.22} \\
    $\gamma{=}2$ & 86.60\,/\,2.32 & 86.80\,/\,2.18 & 87.43\,/\,0.65 & 86.15\,/\,0.53 \\
    \bottomrule
  \end{tabular}
\end{table}
Tables~\ref{tab:sens-wcons-domainnet} and~\ref{tab:sens-wcons-officehome}
report AvgAcc and AvgF on DomainNet and Office-Home as the consolidation
weight $w_{\mathrm{cons}}$ and the drift-weighting sharpness $\gamma$ are
varied. Two trends are consistent. First, under drift-adaptive weighting
($\gamma \ge 1$), increasing $w_{\mathrm{cons}}$ monotonically reduces
forgetting (e.g., $12.66\% \to 4.12\%$ on DomainNet at $\gamma = 2$) while
accuracy remains stable, so the benefit does not hinge on a narrowly tuned
weight. Second, the adaptive weighting is what makes a large preservation
budget safe: with uniform weighting ($\gamma = 0$), the same
$w_{\mathrm{cons}} = 300$ costs $3.6$ accuracy points on DomainNet ($58.25$
vs. $61.80$), whereas concentrating the same budget on high-drift pairs
($\gamma = 2$) attains the lowest forgetting without this accuracy loss.
Office-Home shows the same pattern at a milder scale, with forgetting below
$4.1\%$ over the entire grid.

\section{Training Algorithm}
\label{app:algorithm}
Algorithm~\ref{alg:training} summarizes the full training procedure.

\begin{algorithm}[h]
\caption{Drift-adaptive training over a domain sequence}
\label{alg:training}
\begin{algorithmic}[1]
\Require frozen backbone $f$, encoder $g_\theta$, predictor $q_\phi$, logit
scale $\tau$, EMA teacher $g_\xi$, per-domain class means $\{m_{s,c}\}$ and
context summaries $\{\nu_{s,c}\}$ (initially empty)
\For{domain $t = 1, \dots, T$}
  \If{$t = 1$}
    \State initialize $\mu_c$ with the raw target class-means of
      $\mathcal{D}_1$ from the frozen backbone
  \Else
    \State freeze snapshot $g_{\xi^{(t-1)}}$, $\mu_c^{(t-1)}$
    \State $a_c^{(t-1)} \gets
      \mathrm{normalize}\big(g_{\xi^{(t-1)}}(\mu_c^{(t-1)})\big)$
      \Comment{anchor construction}
  \EndIf
  \For{each mini-batch $(x, y) \sim \mathcal{D}_t$}
    \State pool $u^{\mathrm{raw}}, v^{\mathrm{raw}}$ from CLS-to-patch
      attention \Comment{evidence decomposition}
    \State $u \gets g_\theta(u^{\mathrm{raw}})$;\quad
      $\hat{u} \gets q_\phi\big(g_\theta(v^{\mathrm{raw}})\big)$
    \State compute $\mathcal{L}_{\mathrm{cls}}$ and
      $\mathcal{L}_{\mathrm{pred}}$
    \If{$t \ge 2$}
      \State $r_{s,c} \gets q_\phi\big(g_\theta(\nu_{s,c})\big)$
        for $(s,c) \in \mathcal{S}^{(t)}$ \Comment{replay stored contexts}
      \State $\delta_{s,c} \gets 1 - \cos\big(r_{s,c},\, a_c^{(t-1)}\big)$
        \Comment{drift measurement}
      \State compute $\mathcal{L}_{\mathrm{cons}}$ with drift-adaptive
        weights $\omega_{s,c}$
    \Else
      \State $\mathcal{L}_{\mathrm{cons}} \gets 0$
    \EndIf
    \State update $\theta, \phi, \tau$ by minimizing
      $w_{\mathrm{cls}}\mathcal{L}_{\mathrm{cls}}
       + w_{\mathrm{pred}}\mathcal{L}_{\mathrm{pred}}
       + w_{\mathrm{cons}}\mathcal{L}_{\mathrm{cons}}$;\quad
      $\xi \gets m\,\xi + (1-m)\,\theta$
  \EndFor
  \State store $m_{t,c}$ and $\nu_{t,c}$ for each observed class $c$
  \State recompute $\mu_c^{(t)}$ with drift-weighted combination,
    with $\delta_{s,c}$ evaluated at the end of training on domain $t$
    \Comment{robust prototype construction}
\EndFor
\end{algorithmic}
\end{algorithm}

\begin{figure}[t]
\centering
\includegraphics[width=\textwidth]{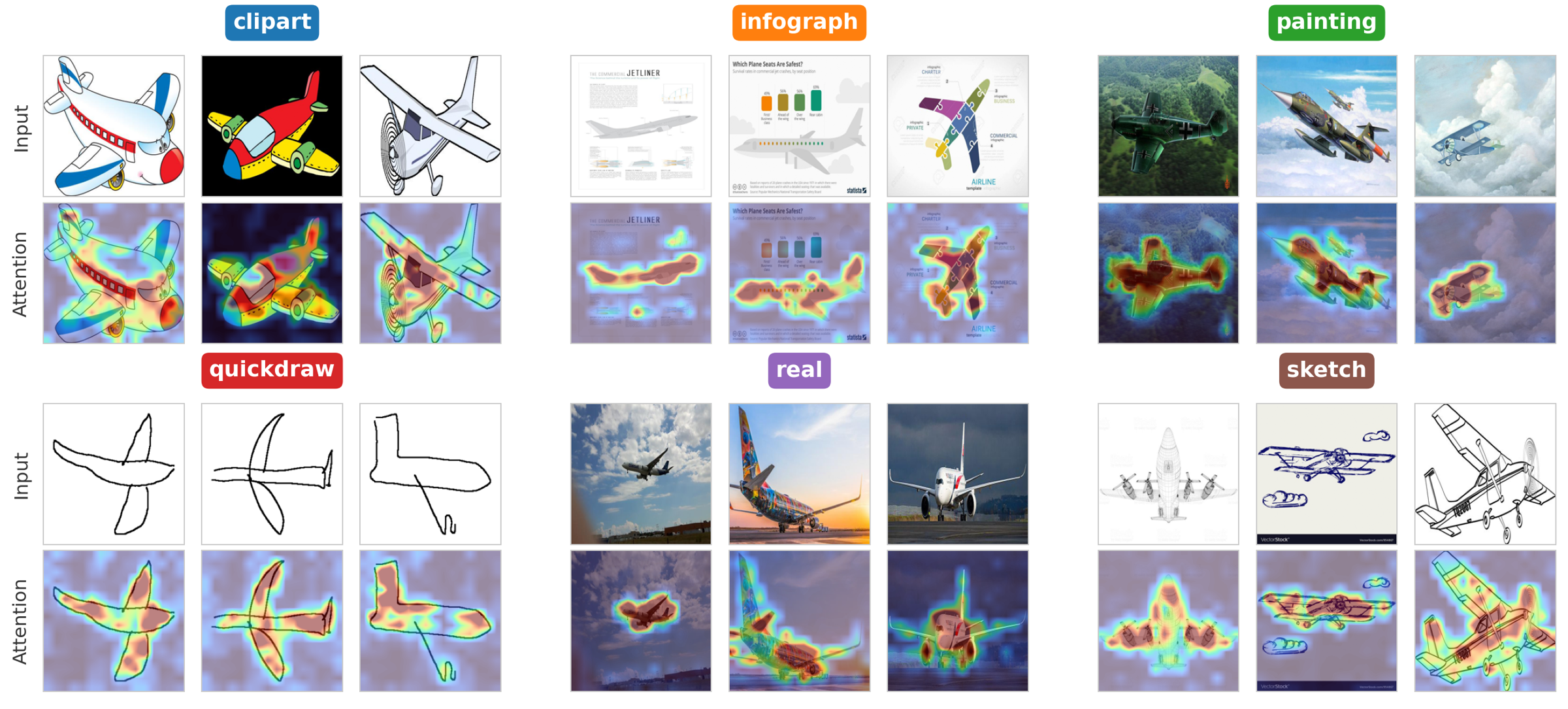}
\caption{CLS-to-patch attention of the frozen backbone for the airplane class
across the six DomainNet domains. Top rows: input images; bottom rows:
attention maps.}
\label{fig:attention}
\end{figure}

\section{Visualization of CLS-to-Patch Attention}
\label{app:attention-vis}
Figure~\ref{fig:attention} visualizes the CLS-to-patch attention $\alpha_n$
of the frozen backbone, which the evidence decomposition of
Sec.~\ref{sec:evidence-prediction} uses to split each image into a target and a context. For the same
class (airplane), we show inputs and attention maps from all six DomainNet
domains. Across domains with sharply different visual statistics---from
photographs (real) to abstract line drawings (quickdraw) and cluttered layouts
(infograph)---the attention concentrates on the object region rather than on
the background or style.

\end{document}